\documentclass[twocolumn]{article}

\usepackage{balance}
\usepackage{amssymb}
\usepackage{multirow}
\usepackage{graphicx}
\usepackage{subcaption}
\usepackage{hyperref}
\usepackage{amsmath}
\usepackage{booktabs}
\usepackage{geometry}
\usepackage{abstract}
\usepackage{natbib}
\usepackage{authblk}
\title{VLM-NCD:Novel Class Discovery with Vision-Based Large Language Models}

\author[1]{Yuetong Su}
\author[2]{Baoguo Wei}
\author[3]{Xinyu Wang}
\author[4]{Xu Li}
\author[5]{Lixin Li}

\affil[1,2,3,4,5]{School of Electronics and Information, Northwestern Polytechnical University, Xi'an, China}
\affil[1]{\texttt{syt15332639458@mail.nwpu.edu.cn}}
\affil[2]{\texttt{wbg@nwpu.edu.cn}}
\affil[3]{\texttt{xw18765@outlook.com}}
\affil[4]{\texttt{lixu@nwpu.edu.cn}}
\affil[5]{\texttt{lilixin@nwpu.edu.cn}}

\begin{document}

\maketitle 

\begin{abstract}
Novel Class Discovery (NCD) aims to utilise prior knowledge of known classes to classify and discover unknown classes from unlabelled data. Existing NCD methods for images primarily rely on visual features, which suffer from limitations such as insufficient feature discriminability and the long-tail distribution of data. We propose LLM-NCD, a multimodal framework that breaks this bottleneck by fusing visual-textual semantics and prototype-guided clustering. Our key innovation lies in (1) modelling cluster centres and semantic prototypes of known classes by jointly optimising known class image and text features, and (2) a dual-phase discovery mechanism that dynamically separates known/novel samples via semantic affinity thresholds and adaptive clustering. Experiments on the CIFAR-100 dataset show that compared to the current state-of-the-art methods, this method achieves up to 25.3\% improvement in accuracy for unknown classes. Notably, our method shows unique resilience to long-tail distributions—a first in NCD literature.
\end{abstract}


\section{Introduction}

With the widespread adoption of open-world intelligent applications, machine vision systems often face the need to identify unknown classes. Novel Class Discovery\cite{han2019learning} aims to automatically identify and classify unknown classes from unlabelled samples using labelled data from known classes, and is one of the core problems in open-world recognition\cite{bendale2015towards}. Existing NCD methods are mainly based on deep visual feature clustering (such as DTC\cite{han2019learning}), but face two major challenges:

\textbf{Insufficient feature discrimination.} Feature extractors trained on known classes in traditional models have difficulty generalising to unknown classes.

\textbf{Long-tail distribution bias.} Rare classes in real data are easily overwhelmed by high-frequency classes, leading to false negatives for novel classes.

In recent years, large-scale visual-language pre-training models (such as CLIP\cite{radford2021learning}) have demonstrated powerful cross-modal semantic understanding capabilities. By training on a vast amount of image-text pairs, these models map images and text into a unified semantic space, making them naturally suited for open-world tasks involving multimodal associations. Currently, CLIP has achieved significant success in visual tasks such as image classification\cite{zhou2022learning,gao2024clip,radford2021learning,zhang2022tip}, object detection\cite{gu2021open,minderer2022simple}, and video understanding\cite{wang2021actionclip}, but its potential for application in the critical field of Novel Class Discovery has not yet been fully explored. This is primarily due to two important reasons:  

\textbf{Insufficient task adaptability.} There is an inherent difference between the fixed class space of closed-set recognition tasks and the open-world characteristics required for Novel Class Discovery. Directly applying CLIP's zero-shot classification paradigm is difficult to adapt to the needs of a dynamically expanding class system.

\textbf{Lack of multimodal collaboration.} Vague descriptions of unknown classes (such as ‘unknown objects’) can easily lead to confusion in clustering boundaries. By fully utilising the complementarity of text and image features, it is possible to effectively develop the unique value of multimodal joint inference in Novel Class Discovery. In particular, the semantic a priori information contained in the text modality can provide key distinguishing clues for unknown classes.

For novel class discovery in open-world scenarios, we should leverage all available background knowledge or prior information beyond just the given labeled samples. Additional relevant knowledge can enhance classification confidence and thus improve model's novel class discovery capability. In this work, we employ Large Language Models as the representation of such background knowledge. Therefore, to address the above issues, this paper proposes a Novel Class Discovery method based on Large Language Models , as shown in Figure 1. We propose a Novel Class Discovery framework based on multimodal semantic alignment. The core idea is to enhance class distinguishability through image-text feature fusion and to alleviate long-tail bias using a text-guided dynamic allocation strategy. 

In the top section, we first utilise CLIP's aligned visual-language representations, using the input image as a query to retrieve a set of highly relevant text descriptions from a large knowledge databases. To further leverage CLIP's large-scale pre-trained representations, the input image and its retrieved text are encoded into a set of feature encodings using a frozen CLIP image and text encoder. In the bottom section, given the connected text and image views, we employ a semi-supervised k-means clustering method to cluster the features into known and unknown classes. The specific contributions are as follows:

\begin{figure}[t]
\centering
\includegraphics[width=0.9\columnwidth]{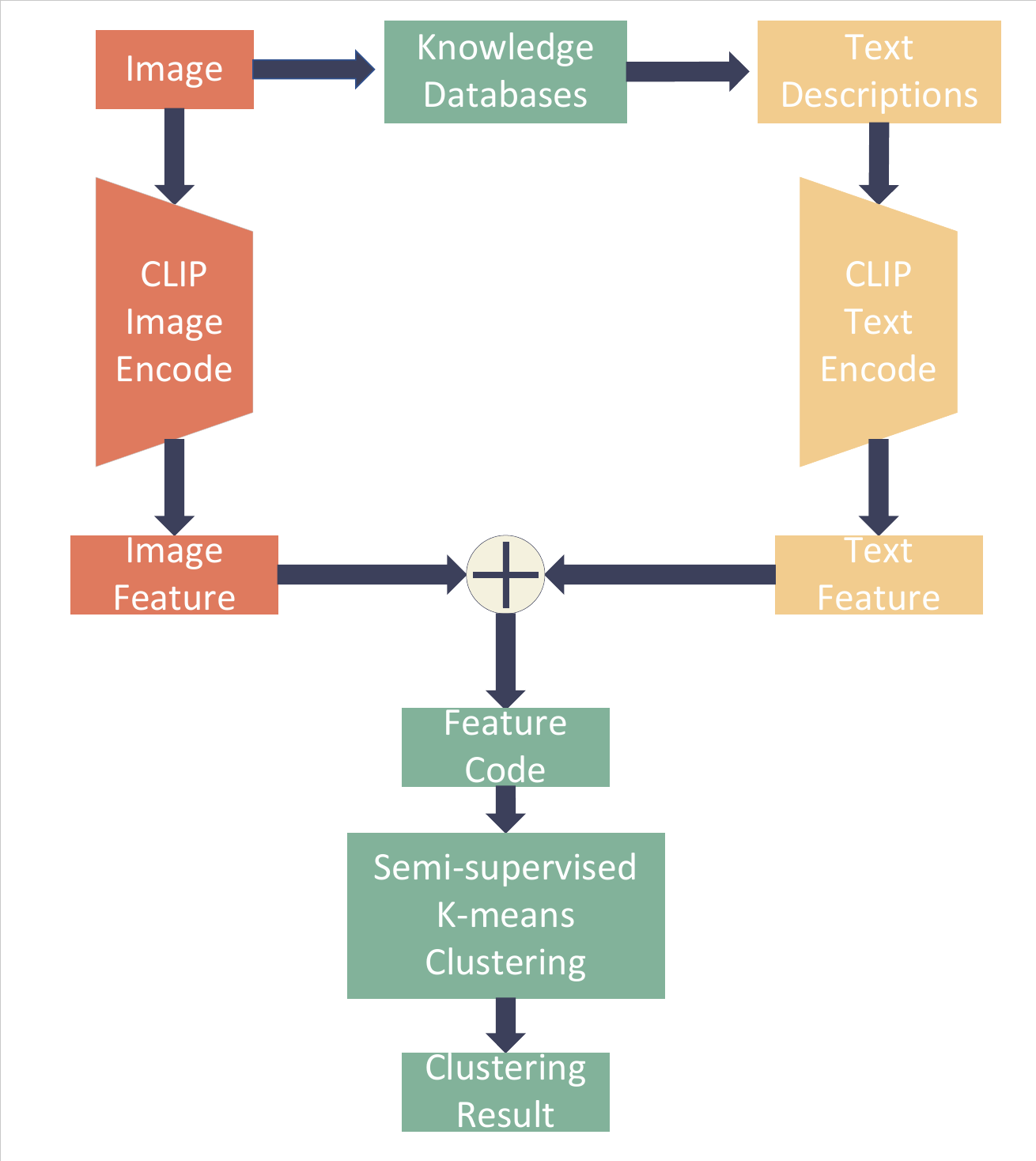}
\caption{Methodology of this paper.}
\label{fig1}
\end{figure}

\textbf{Multimodal feature fusion mechanism.} We propose the first multimodal NCD framework based on CLIP, which utilises joint image and text features to enhance class distinguishability. The image and text features of CLIP are concatenated to generate a joint representation, while retaining visual granularity information and textual semantic constraints, thereby improving the discriminative power for known and unknown classes.

\textbf{Semantically guided two-stage discovery.} Design a semantically guided two-stage discovery mechanism that dynamically separates known classes from novel classes based on text similarity. In the known class modelling stage, construct text semantic prototypes. In the NCD stage, first filter known class samples based on text similarity, then perform k-means clustering on the remaining samples to reduce semantic drift.

\textbf{Lightweight solution.} High performance can be achieved with only a pre-trained CLIP model, without the need for fine-tuning or complex training processes, providing a lightweight solution for open-world recognition.

Experiments show that this method outperforms existing methods with novel class clustering accuracy rates of 85.6\% and 78.2\% on the CIFAR-100 and ImageNet-100 benchmarks, respectively. Ablation experiments validate the critical role of Large Language Models as feature extractors and text-guided strategies for long-tail data.


\section{Related work}

\subsection{Novel Class Discovery}

Novel Class Discovery (NCD) is a relatively emerging field, initially proposed as ‘cross-task transfer’, which involves transferring learning from labelled data to clustering unseen classes in unlabelled data. As an important research direction in open-world recognition, it has made significant progress in recent years. \cite{hsu2017learning,hsu2019multi} use a paired conjoint network trained on labelled data and applies it to a clustering network trained on unlabelled data. Subsequent research\cite{han2019learning} improved upon this approach using specialized deep clustering methods. In RankStat[\cite{han2020automatically,han2021autonovel}, a three-stage pipeline was deployed: first, the model was pre-trained on all data for representation learning, then fine-tuned on labelled data to capture higher-level semantic knowledge, and finally, ranking statistics were used to transfer knowledge from labelled to unlabelled data. \cite{zhong2021neighborhood} proposed a contrastive learning method that generates hard negatives by mixing labelled and unlabelled data in the latent space. UNO\cite{fini2021unified} introduced a unified cross-entropy loss that jointly trains the model on labelled and unlabelled data by swapping pseudo-labels in the classification head. The original GCD\cite{vaze2022generalized} method involves k-means clustering of DINO embeddings, while XCon\cite{fei2022xcon} improves these results through additional contrastive training.

However, these methods have two main limitations. First, they typically train feature extractors on limited known class data, resulting in insufficient generalisation ability for unknown classes; second, these methods fail to effectively utilise semantic association information between classes, resulting in poor performance when processing data with complex semantic structures. Our work is based on a more realistic new environment\cite{vaze2022generalized}, where unlabelled samples can come from either known or unknown classes. In our method, we focus on utilising multimodal models in multiple ways, and we demonstrate superior results compared to all previously published techniques.

\subsection{Large Language Models}

In recent years, large-scale pre-trained models have demonstrated tremendous potential in the field of computer vision. The CLIP\cite{radford2021learning} model was pre-trained on 400 million image-text pairs, establishing cross-modal semantic alignment capabilities. This contrastive learning-based training method enables the model to achieve outstanding zero-shot transferability. In image classification tasks, Gao et al.\cite{gao2024clip} validated CLIP's performance in few-shot classification tasks, but their core experiments directly froze the CLIP visual encoder (without fine-tuning) and only adapted features through a simple linear layer. Radford et al.\cite{radford2021learning} incorporated the CLIP visual and text encoders into a classifier based on image-text similarity; Zhang et al.\cite{zhang2022tip} froze the CLIP visual encoder, and constructing a classifier using cached features, achieving a method that successfully classifies few-shot data without training. In object detection tasks, Gu et al.\cite{gu2021open} replaced the detector backbone with the CLIP visual encoder and distilled textual knowledge into the detection head to achieve open-word detection; Minderer et al.\cite{minderer2022simple} designed a text-conditioned detection framework based on the CLIP visual encoder to achieve open-word detection. In segmentation tasks, Xu et al.\cite{xu2022groupvit} extended the CLIP visual encoder and successfully achieved zero-shot semantic segmentation through a text-supervised learning grouping mechanism.

However, the application potential of CLIP in the critical field of Novel Class Discovery has not yet been fully explored. In our method, we not only fully leverage the capabilities of the CLIP visual-language large model but also simultaneously utilise both the visual encoder and text encoder, successfully validating CLIP's outstanding capabilities in the field of Novel Class Discovery and laying a solid foundation for the use of other large models.

\subsection{Image-text multimodal}

Multimodal learning offers new approaches for NCD tasks. \cite{han2021autonovel} is the first to use image-text embedding to discover novel classes, and combining visual and linguistic features can significantly improve model performance. In the field of image classification, CLIP\cite{radford2021learning} aligns image and text embeddings through contrastive learning, enabling zero-shot classification; Jia et al.\cite{jia2021scaling} trained multimodal models using noisy network data to enhance classification generalisation capabilities. In the field of object detection, Li et al.\cite{li2022grounded} jointly trained detection and language models to achieve semantically guided object detection; Zareian et al.\cite{gu2021open} utilised CLIP knowledge distillation to achieve open-word detection. In the field of object tracking, Yang et al.\cite{yang2020grounding} combined language descriptions with visual features for object tracking; Sha et al.\cite{sha2023large} focused on autonomous driving tasks and utilised language models to enhance tracking semantic understanding.

However, in NCD tasks, the issues of insufficient feature discriminability and the long-tail distribution of data severely limit feature extraction capabilities for novel classes. To address this challenge, this paper proposes a solution that combines visual and semantic information. By integrating image features with class text knowledge, this approach effectively enhances the feature representation capabilities for novel classes. Specifically, the method leverages the semantic prior information provided by the text modality to supplement the features of scarce samples while also providing additional supervisory signals for tail classes under long-tail distributions, thereby improving the model's generalisation performance.


\section{Method}

In this section, we first introduce the symbols and definitions of NCD. Then, we explain how to use CLIP in NCD and introduce our method for handling this task.

\subsection{NCD Problem setting}

The dataset $D$ consists of a labelled subset of dataset $D_{\mathcal{L}} = \{(X_i, y_i)\}_{i=1}^N \in \mathcal{X} \times \mathcal{Y}_{\mathcal{L}}$ and an unlabelled dataset $D_u = \{(X_i, Y_i)\}_{i=1}^M \subseteq \mathcal{X} \times \mathcal{Y}_u$, where $\mathcal{Y}_{\mathcal{L}} \subseteq \mathcal{Y}_u$. The goal is to learn a model that can group $D_u$ instances based on  information from the $D_{\mathcal{L}}$ dataset. Leveraging advancements in visual transformers and their significant performance in self-supervised learning\cite{caron2021emerging}, Vaze et al. \cite{vaze2022generalized} designed a two-stage training process for the NCD task: supervised contrastive learning on labelled data and unsupervised contrastive learning on all data.

Let $X_i$ and $X_i'$ be two randomly enhanced views of the same image in a small batch $B$. The unsupervised contrast loss is defined as follows:

\[
\mathcal{L}_i^u = -\log \frac{\exp(Z_i \cdot Z_i' / \tau)}{\sum_n \mathbb{1}_{[n \neq i]} \exp(Z_i \cdot Z_n' / \tau)'}
\]

where $Z_i = h(f(x_i))$ is the feature of the input image $x_i$ extracted by the backbone network $f(\cdot)$ and projected onto the embedding space through the projection head $h(\cdot)$ , and $Z_i'$ is the feature of another view of the input image $x_i$.

The definition of supervised contrast loss is as follows:

\[
L_i^s = -\frac{1}{|N(i)|} \sum_{q \in N(i)} \log \frac{\exp(Z_i \cdot Z_q / \tau)}{\sum_n \mathbb{1}_{[n \neq i]} \exp(Z_i \cdot Z_n / \tau)'}
\]

where $N(i)$ indicates the index of other images with the same label as $X_i$ in the small batch $B$ . Then, the final objective function is a combination of these two losses:

\[
\mathcal{L}^t = (1-\lambda) \sum_{i \in \mathcal{B}_\mathcal{L} \cup \mathcal{B}_{\mathcal{U}}} \mathcal{L}_i^u + \lambda \sum_{i \in \mathcal{B}_\mathcal{L}} \mathcal{L}_i^s
\]

where $\lambda$ is a weighting factor, $\mathcal{B}_\mathcal{L}$ and $\mathcal{B}_{\mathcal{U}}$ represent small batches of labelled and unlabelled images, respectively. For label assignment, the authors propose a semi-supervised k-means algorithm, which is similar to k-means\cite{mcqueen1967some} but with an important difference: when calculating the clustering assignment at each step, the semi-supervised k-means algorithm considers the labelled data $\mathcal{D}_\mathcal{L}$. This means that labelled samples will always be assigned to the correct cluster, regardless of their distance from the nearest cluster centre.

\subsection{Our method}

By combining text and image information, language-image models can achieve better performance in various tasks. Therefore, we suggested utilising CLIP's zero-shot learning capabilities and multimodal alignment encoders to address two major issues in NCD, and propose a retrieval-based data augmentation method, with the model architecture shown in Figure 2.

\begin{figure*}[t]
\centering
\includegraphics[width=0.8\textwidth]{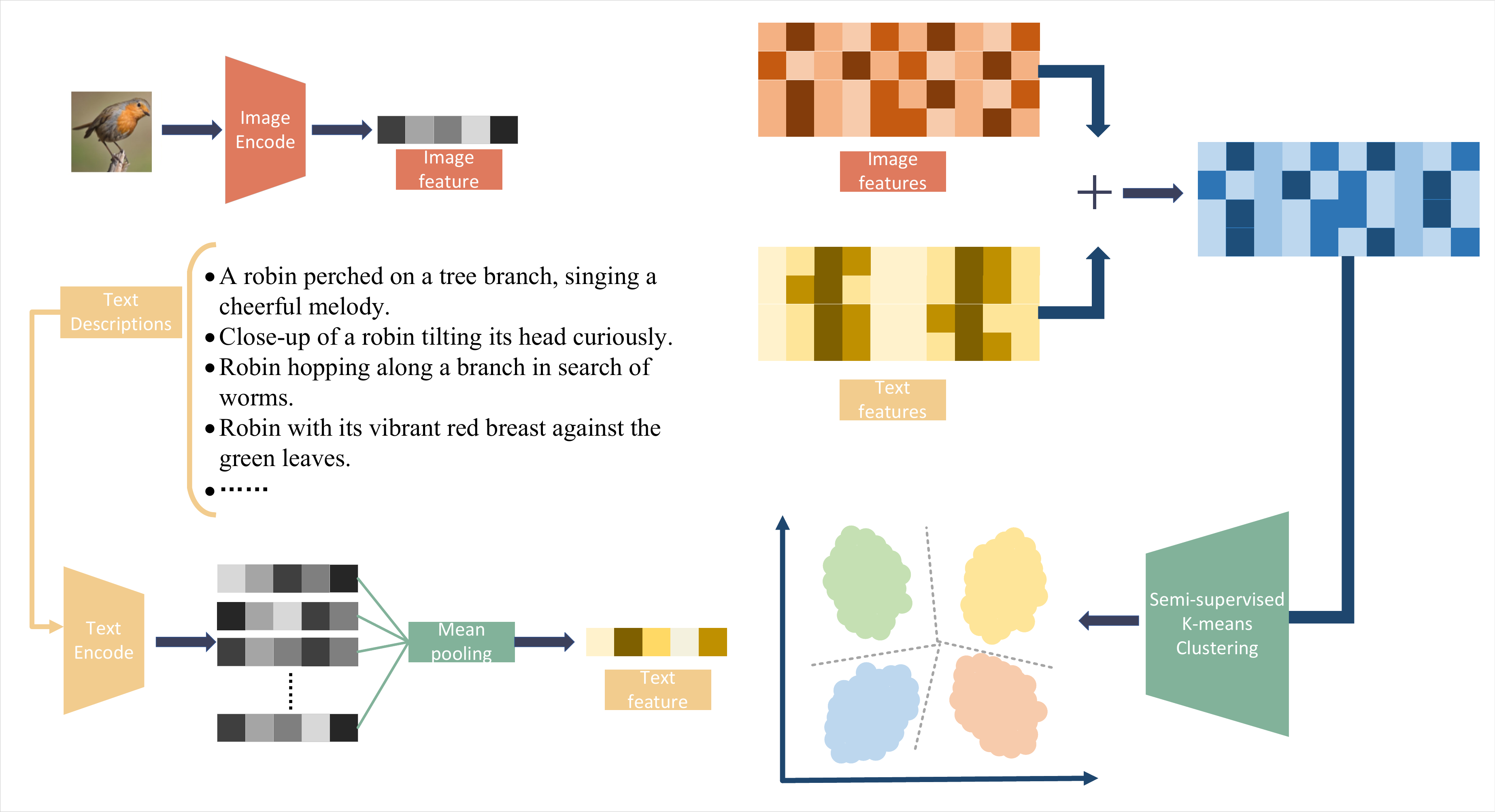}
\caption{Model architecture. First, we propose a cross-modal retrieval module to retrieve context-aware text descriptions as supplementary information for clustering, complementing labelled and unlabelled data. Second, we concatenate image views and text views and use a semi-supervised k-means clustering algorithm to group known and unknown classes.}
\label{fig2}
\end{figure*}

\subsubsection{CLIP in NCD}

We address the NCD task by leveraging CLIP's multimodal joint embeddings. The CLIP model has two branches, the image branch CLIP-Image and the text branch CLIP-Text, which encode images and text into global feature representations, respectively. CLIP is trained on a large-scale image-text pair dataset, ensuring that paired images and texts are close in the embedding space, while unpaired ones are pushed apart. 

To improve data representation, especially for labelled and unlabelled data, we optimise representations by combining supervised contrastive learning on labelled data and unsupervised contrastive learning on all data. We achieve this by fine-tuning the model on the target data simultaneously. CLIP learns image representations by comparing the representation of an image with its textual description, such as ‘a photo of {class name}’. The text description is called a prompt, and its design is crucial for improving CLIP's performance. 

However, the unlabelled data contains latent novel classes that are entirely unknown — we possess no prior knowledge (e.g., class names, quantity, or semantic descriptions) to even provisionally define them, rendering prompt-based methods fundamentally inapplicable due to the absence of a valid reference for prompt construction. Therefore, inspired by research in the field of image caption generation, we generate a set of text descriptions for both labelled and unlabelled data, providing supplementary information for the input image, as shown in Figure 3, which includes details and information about the input image to map it to the feature space. Training a separate caption generation model to generate text descriptions may be expensive and challenging, so for each labelled and unlabelled image, we retrieve the k most relevant descriptions from a text corpus. Current mainstream description databases include Conceptual Captions (3M)\cite{sharma2018conceptual}, Conceptual Captions (12M)\cite{changpinyo2021conceptual}, MS Coco\cite{lin2014microsoft}, and LION\cite{schuhmann2021laion}, among others.

\begin{figure}[t]
\centering
\includegraphics[width=0.9\columnwidth]{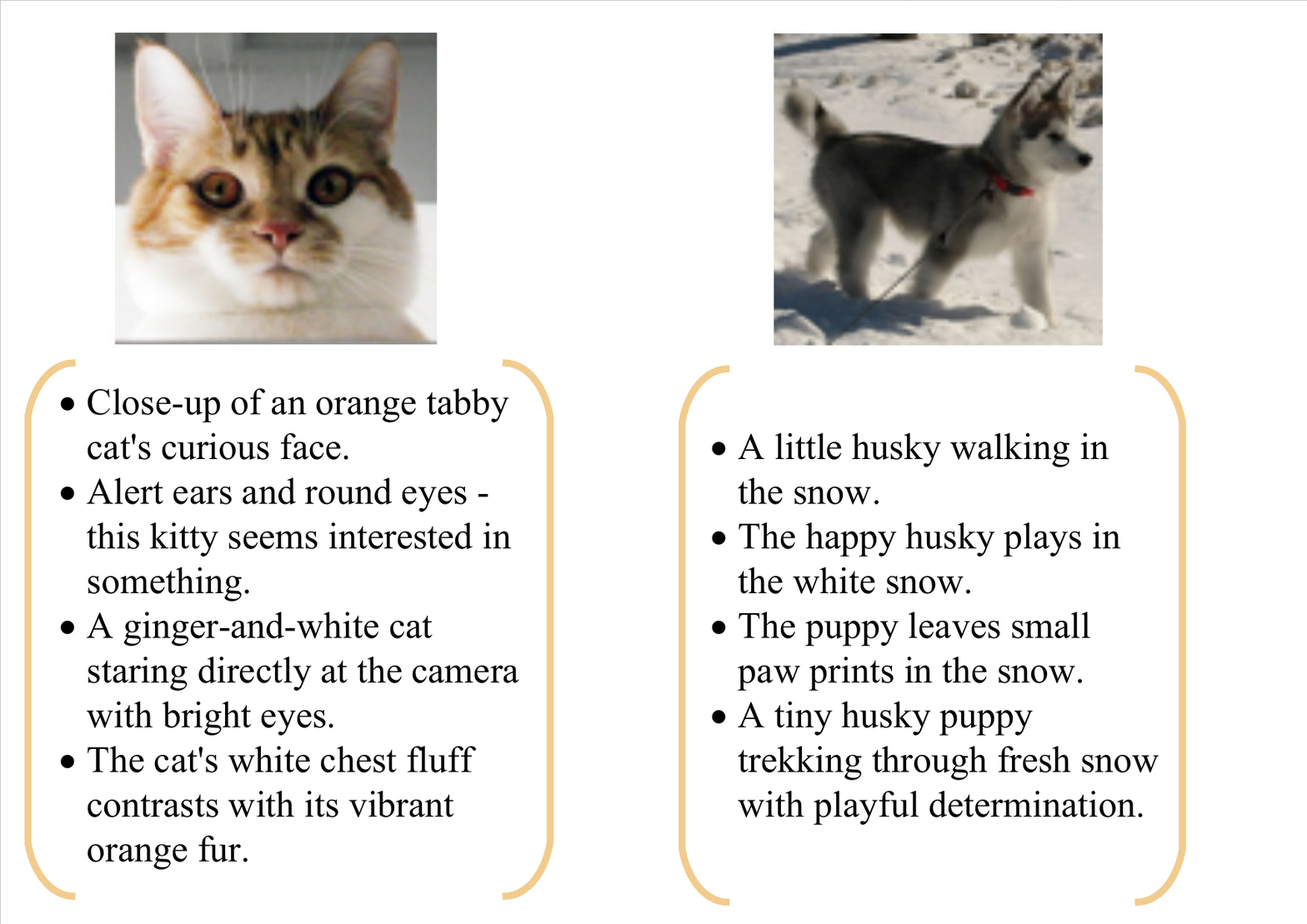}
\caption{Shows the top 4 text descriptions most relevant to an image in the ImageNet dataset, retrieved from the Conceptual Captions (3M) database.}
\label{fig3}
\end{figure}

\textbf{Text description retrieval.} Given an image query, the goal is to retrieve the top k most relevant text descriptions from the description database. To achieve this goal, we propose using cross-modal joint embedding from CLIP\cite{radford2021learning} for cross-modal retrieval tasks. Specifically, we use CLIP-Text to encode all descriptions in the description database as search terms. Images are encoded as queries by CLIP-Image. We then search the description database for the top k text descriptions with the highest cosine similarity scores. Examples of the top 4 results are shown in Figure 3.

\textbf{Multi-view generation.} Our feature vector extraction and multi-view generation framework is illustrated in Figure 2 (left). Given an image and a set of text descriptions, we first encode the image using the CLIP image encoder, then encode the text descriptions using the CLIP text encoder, and generate a view (sentence embedding) through mean pooling. Finally, the feature vectors of the image and text views are concatenated and projected into the CLIP latent space, where clustering is performed directly.

\textbf{Semi-supervised k-means clustering.} Given image views and text views, we concatenate the feature vectors and apply semi-supervised k-means clustering according to the method in \cite{vaze2022generalized} to group unlabelled data into known and unknown classes. Semi-supervised k-means clustering is a constrained version of the traditional k-means method, where the number of clusters k is assumed to be known. This requires $\mathcal{D}_\mathcal{L}$ data instances to be assigned to the correct cluster centres based on their true class labels. First, use real class labels to obtain $\mathcal{D}_\mathcal{L}$ data in semi-supervised k-means clustering centres. Second, for novel classes, use k-means++\cite{arthur2006k} to obtain cluster centres from the data. During the process of updating and assigning cluster centres, instances of the same class are always grouped together, while instances can be assigned to any cluster based on their distance to each cluster centre. Once the algorithm converges, each instance can be assigned a cluster label.


\section{Experiment}

\subsection{Model Architecture Details}

CLIP\cite{radford2021learning} consists of two encoders, CLIP-Image and CLIP-Text, which are pre-trained Transformer models for images and text, respectively. CLIP-Text is a basic Transformer model comprising 12 layers with a hidden size of 768, and the final linear projection layer generates a representation vector of size 512. CLIP-Image is a hybrid ViT-Base model (the same as the DINO training model, for fair comparison), consisting of 12 stacked layers, starting with a convolutional layer for feature extraction. For a given image, 49 embeddings with a hidden size of 768 are generated, and the output hidden state is projected from 768 to 512 dimensions to match the output of the CLIP-Text encoder. We fine-tuned the last block of the visual transformer with an initial learning rate of $5e^{-5}$ and decayed it over time using cosine annealing scheduling. We trained the model for 100 epochs using batches of size 128 and set the $\lambda$ value in the loss function to 0.25. The optimal hyperparameters were selected by tuning and testing on a separate validation set.

\subsection{Datasets and Evaluation metrics}

\textbf{Datasets.} We evaluate our method on general image classification tasks. Following the approach outlined in \cite{vaze2022generalized}, we selected CIFAR-10, CIFAR-100, and ImageNet-100 as general image classification datasets. We divided the training data into labelled datasets and unlabelled datasets. We equally divided all classes into known classes and unknown classes, then randomly sampled 50\% of the images from the known classes as unlabelled data to ensure that the unlabelled dataset $\mathcal{D}_\mathcal{u}$ contains images from both known and unknown classes, while the labelled dataset only contains images from known classes. These divisions are summarised in Table 1.

\textbf{Evaluation metrics.} We use clustering accuracy (ACC) to measure the performance of the model, which is defined as follows:

\[
\mathrm{ACC} = \max_{p \in P({y}_U)} \frac{1}{N} \sum_{i=1}^N \mathbb{1}\{y_i = p(\hat{y}_i)\}
\]

where ACC is the clustering accuracy, which is calculated using the Hungarian method [32] to match the model's predictions $\hat{y}_i$ with the true labels $y_i$. N is the total number of images in the unlabelled dataset. Following the approach in \cite{vaze2022generalized}, we use this metric on three different sets, namely:

\textbf{All.} Refers to the entire unlabeled set $\mathcal{D}_u$.
\textbf{Old.} Refers to images in the unlabelled dataset $\mathcal{D}_u$ that belong to $\mathcal{Y}_\mathcal{L}$.
\textbf{New.} Refers to images in the unlabelled dataset $\mathcal{D}_u$ that belong to $y_u \setminus y_\mathcal{L}$.

\begin{table}[t]
\centering
\begin{tabular}{llll}
\hline
      & CIFAR-10 & CIFAR-100 & ImageNet-100 \\
\hline
    $|y_\mathcal{L}|$ & 5 & 80 & 50 \\
    $|y_u|$ & 10 & 100 & 100 \\
    $|\mathcal{D}_\mathcal{L}|$ & 12.5k & 20k & 31.9k \\
    $|\mathcal{D}_u|$ & 37.5k & 30k & 95.3k \\
\hline
\end{tabular}
\caption{Dataset division. $(|y_\mathcal{L}|, |y_u|)$ represents the number of classes in the labelled and unlabelled sets. $(|\mathcal{D}_\mathcal{L}|, |\mathcal{D}_u|)$ represents the number of images in each set.}
\label{table1}
\end{table}

\subsection{Comparison results}

Our method is first compared with leading techniques methods on image classification. RankStats+\cite{han2021autonovel} and UNO+\cite{fini2021unified} are two competitive baseline methods for NCD, modified for the GCD setting. XCon\cite{fei2022xcon} is a method tailored for the GCD setting. The results in the General Image Recognition benchmark are shown in Table 2. Across all datasets in our experiments, our method achieves the best performance on most classes, often improving significantly over previous methods. On ImageNet-100 and CIFAR-100, our method outperforms other methods on all subsets, further confirming that the dual use of multimodal models enhances performance compared to using visual models alone.

\begin{table*}[t]
\centering
\begin{tabular}{llllllllll}
\hline
      & \multicolumn{3}{c}{CIFAR-10} & \multicolumn{3}{c}{CIFAR-100} & \multicolumn{3}{c}{ImageNet-100} \\
\hline
    Class & All & Old & New & All & Old & New & All & Old & New \\
\hline
    RankStats+\cite{han2021autonovel} & 46.8 & 19.2 & 60.5 & 58.2 & 77.6 & 19.3 & 37.1 & 61.6 & 24.8 \\
    UNO+\cite{fini2021unified} & 68.6 & \textbf{98.3} & 53.8 & 69.5 & 80.6 & 47.2 & 70.3 & 95.0 & 57.9 \\
    GCD\cite{vaze2022generalized} & 91.5 & 97.9 & 88.2 & 73.0 & 76.2 & 66.5 & 74.1 & 89.8 & 66.3 \\
    XCon\cite{fei2022xcon} & 96.0 & 97.3 & 95.4 & 74.2 & 81.2 & 60.3 & 77.6 & 93.5 & 69.7 \\
    Ours & \textbf{96.6} & 97.2 & \textbf{96.4} & \textbf{85.2} & \textbf{85.0} & \textbf{85.6} & \textbf{84.0} & \textbf{95.5} & \textbf{78.2} \\
\hline
\end{tabular}
\caption{Comparison results of general image recognition datasets.}
\label{table2}
\end{table*}

\begin{table*}[t]
\centering
\begin{tabular}{llllll}
\hline
    Datasets & Image Encoders & Knowledge Databases & All & Old & New \\
\hline
    \multirow{4}{*}{CIFAR-100}  & DINO & N & 73.0 & 76.2 & 66.5 \\
           & DINO & Y & 75.9 & 79.7 & 67.3 \\
           & CLIP & N & 84.2 & 83.1 & 82.3 \\
           & CLIP & Y & \textbf{85.2} & \textbf{85.0} & \textbf{85.6} \\
\hline
\end{tabular}
\caption{Clustering accuracy on whether using knowledge datasets and different image encoders.}
\label{table3}
\end{table*}

\begin{figure*}[t!]
\centering
  \begin{subfigure}[b]{0.48\textwidth}
    \includegraphics[width=\linewidth]{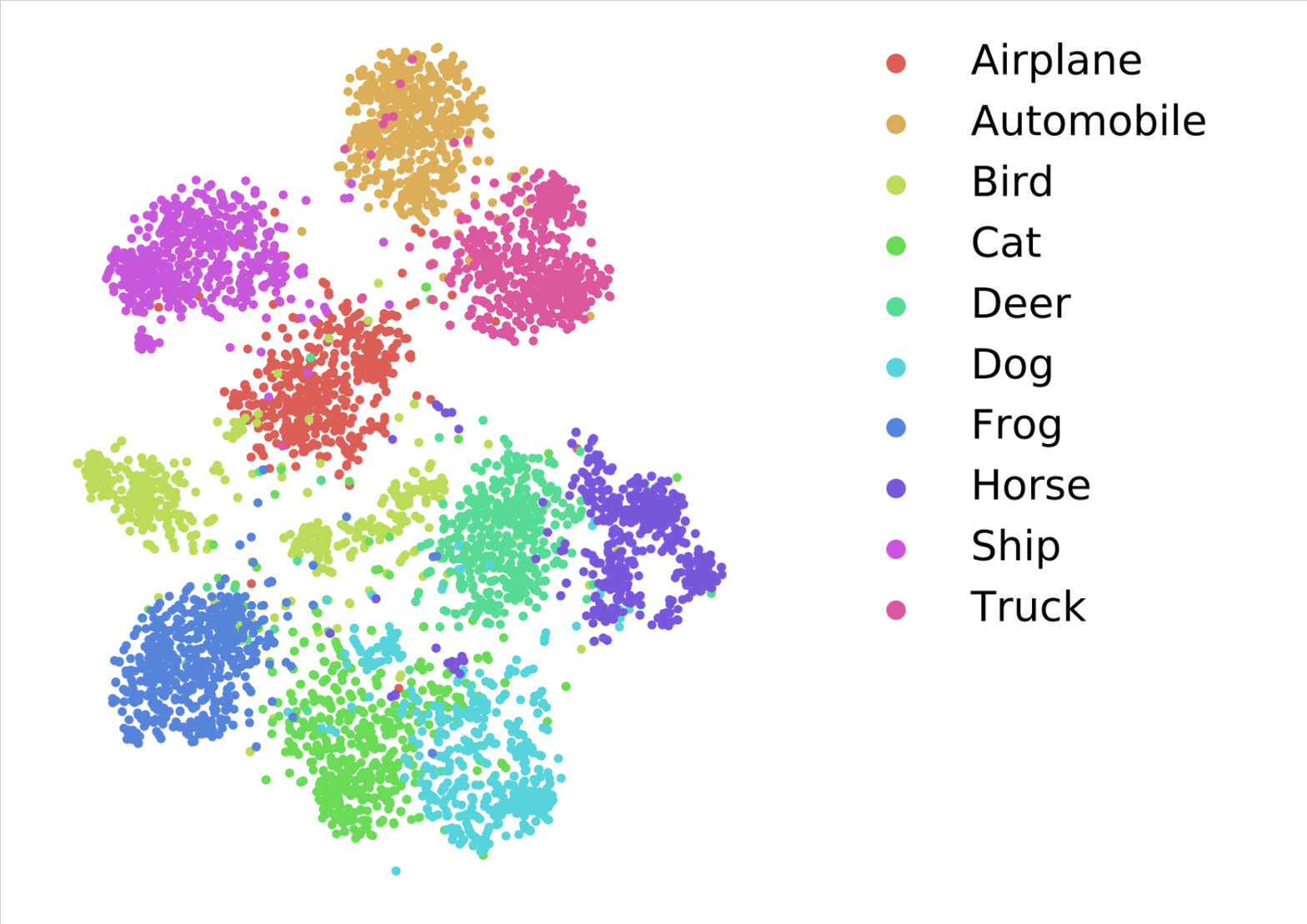}
    \caption{t-SNE results using only image features on the CIFAR-10 dataset}
    \label{fig:left}
  \end{subfigure}
  \hfill 
  \begin{subfigure}[b]{0.48\textwidth}
    \includegraphics[width=\linewidth]{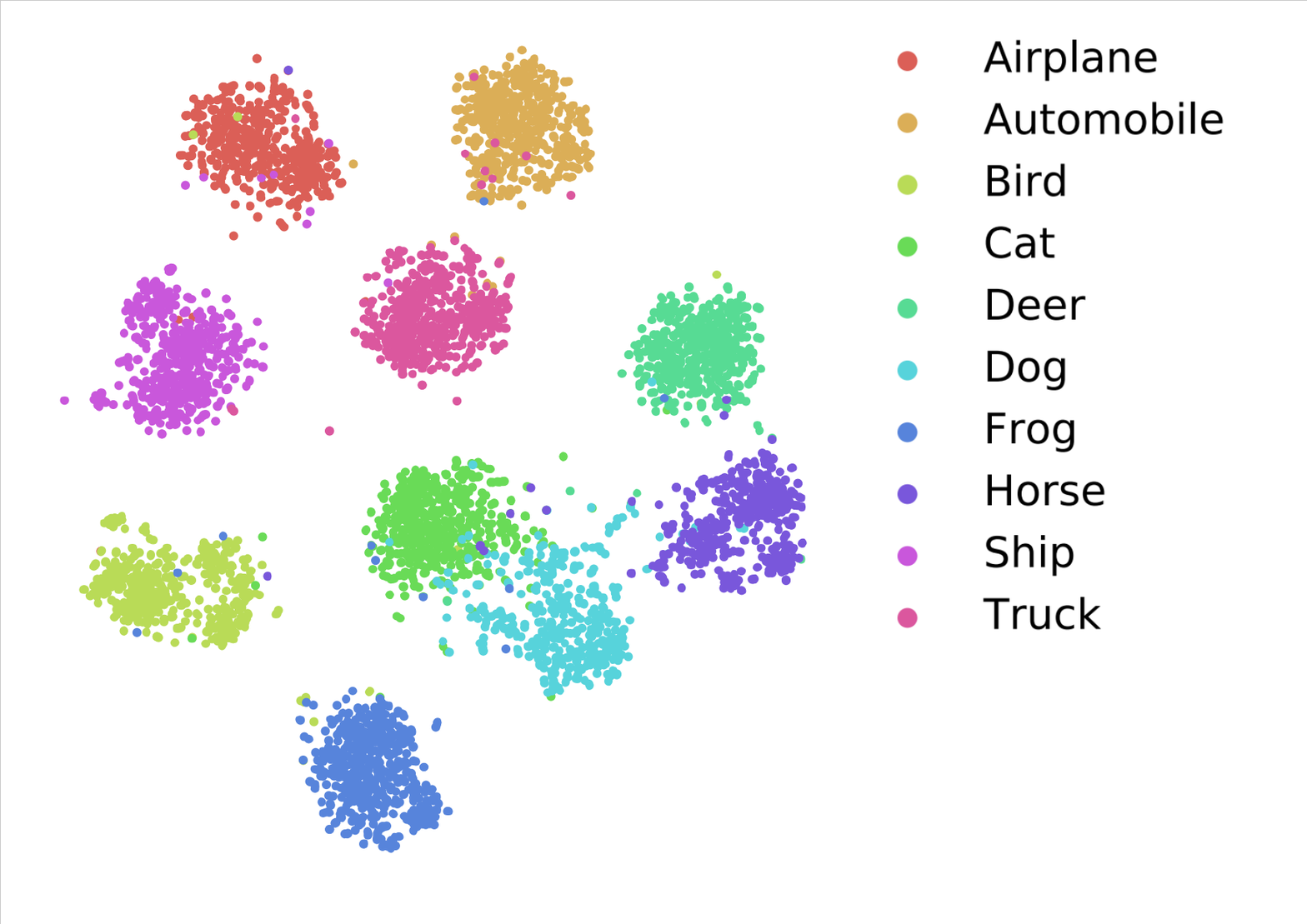}
    \caption{t-SNE results using both image and text features on the CIFAR-10 dataset}
    \label{fig:right}
  \end{subfigure}
\caption{t-SNE results showing whether text features were used in the CIFAR-10 dataset.}
\label{fig4}
\end{figure*}

\begin{table}[t]
\centering
\begin{tabular}{lllll}
\hline
    Datasets & Knowledge Databases & All & Old & New \\
\hline
    \multirow{5}{*}{CIFAR-100}  & CC-12M & \textbf{85.2} & 85.0 & \textbf{85.6} \\
           & CC-3M & 82.8 & 82.6 & 83.2 \\
           & MSCOCO & 85.1 & \textbf{85.5} & 84.2 \\
           & LAION-400M & 82.0 & 82.6 & 80.8 \\
           & LAION-5B & 82.5 & 83.4 & 80.6 \\
\hline
\end{tabular}
\caption{Accuracy of models using different knowledge databases as text description sources.}
\label{table4}
\end{table}

\begin{figure}[t]
\centering
\includegraphics[width=0.9\columnwidth]{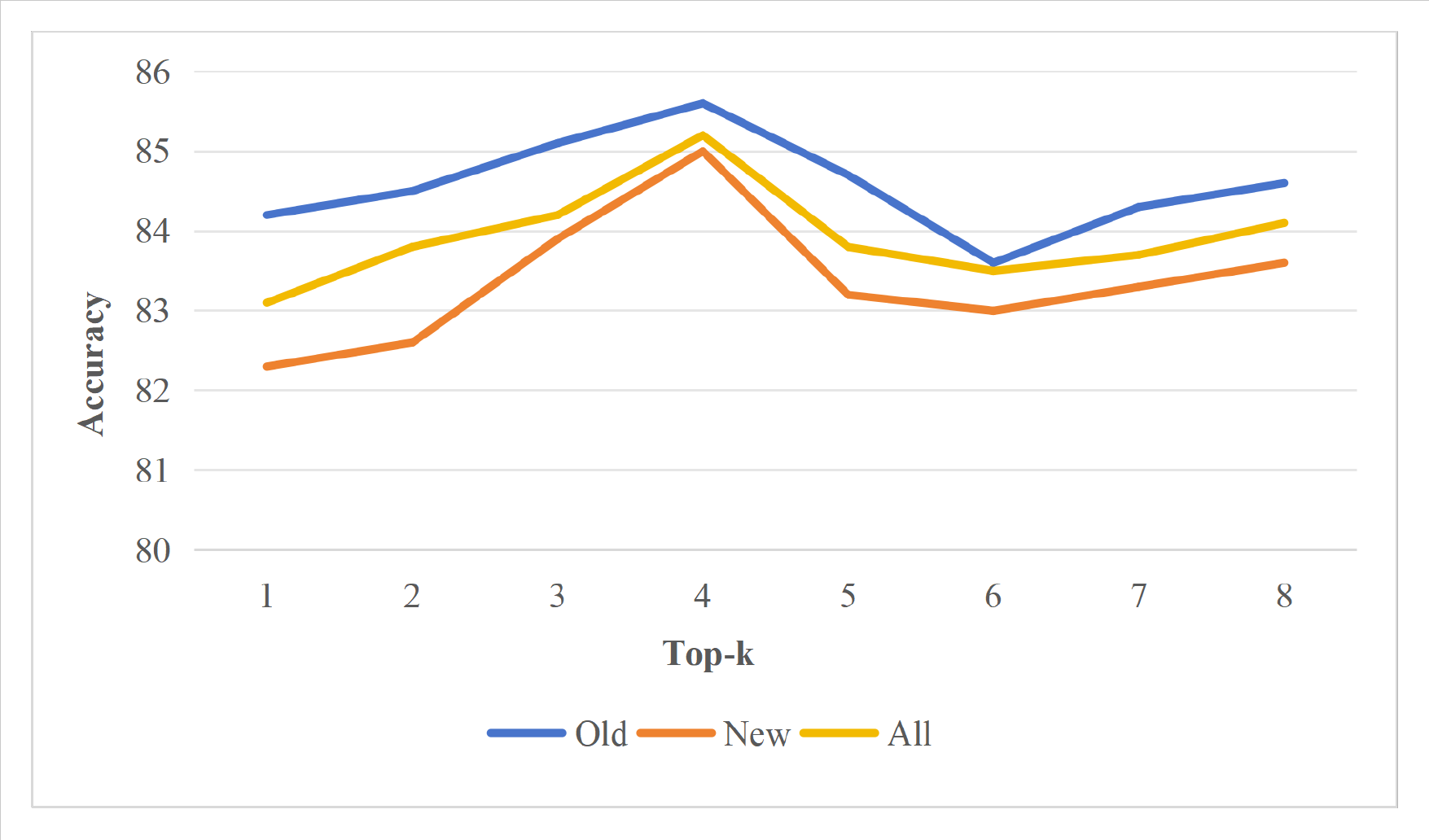}
\caption{The impact of different Top-k values on the results.}
\label{fig5}
\end{figure}

\subsection{Analysis}

We analysed the contribution of certain aspects of our method through ablation experiments. Specifically, we emphasised the importance of the impact of different image encoders, the impact of whether to use text descriptions and different description databases, and the impact of the number of texts k for each image on model accuracy.

As shown in Table 3, the table compares the clustering accuracy of models using different image encoders and whether they use only visual features or both visual and text features. DINO refers to the use of pre-trained weights from GCD's DINO\cite{caron2021emerging}, while CLIP refers to the use of a pre-trained ViT-B/16 backbone network. The results show that models using both image and text features with CLIP significantly outperform those using only image features, confirming that incorporating language information can significantly improve performance compared to image-only models. Additionally, the retrieval mechanism, which uses CLIP's cross-modal joint embeddings for text retrieval, further enhances performance.

Overall, in NCD settings, semantic information plays a crucial role in improving model representation and performance, and retrieval-based augmentation mechanisms further improve clustering accuracy compared to using visual features alone. Descriptive retrieval text is essential to our method. In typical datasets, text descriptions may vary in their association with images. Ideally, we aim to encode salient objects and meaningful details within images to enhance representation learning for object recognition tasks.

For contrastive models, the learned representations are influenced by the text encoder, suggesting that textual descriptions of image scene content can enhance the transferability of CLIP models. We validated this hypothesis and quantified the descriptiveness of textual descriptions using multiple text datasets. We conducted the first four cross-modal retrieval experiments on multiple data sources, including Conceptual Captions (3M)\cite{sharma2018conceptual}, Conceptual Captions (12M)\cite{changpinyo2021conceptual}, MS Coco\cite{lin2014microsoft}, and LION\cite{schuhmann2021laion}, and recorded the model's accuracy on All, Old, and New subsets for each dataset. Table 4 shows the model's results on the CIFAR100 dataset.

\subsection{Qualitative results}

We further demonstrate the results of t-SNE\cite{kuhn1955hungarian} projections using only image features and simultaneously using image and text features on the CIFAR-10 dataset, as shown in Figure 4, where we present the clustering characteristics of unlabelled data. When using only image features, data points of the same class are typically projected near each other, forming distinct clusters, but there is some overlap between classes. In contrast, when using both image and text features, the image-text features form distinct clusters with more pronounced separation, further confirming the utility of language in this setting.

In addition, we examined the model's sensitivity to the number of images retrieved (top-k) and demonstrated its accuracy using Conceptual Captions (12M)\cite{changpinyo2021conceptual} on the CIFAR-100 dataset. Figure 5 shows that some text descriptions may not contain useful information, causing the model's accuracy to plateau or even decline after a certain number.


\section{Conclusion}

In this paper, we propose a new Large Language Models-based approach to address the Novel Class Discovery problem, leveraging the multimodal capabilities of the CLIP model. First, we utilise the image encoder of CLIP to establish a robust baseline for NCD. Second, we introduce a supplementary retrieval-based augmentation method, specifically retrieving text descriptions from a text corpus and jointly clustering image and text embeddings. We conducted a rigorous analysis to demonstrate that our method is highly suitable for the NCD setting.

We present quantitative results on three general classification datasets, achieving significant performance improvements over previous methods. Additionally, we demonstrate that the two components of CLIP are complementary and necessary, working together to achieve strong performance. However, there are still some limitations and future work, including improving the retrieval process to enhance the quality of retrieved contextual knowledge and conducting experimental validation on other fine-grained datasets.





\bibliographystyle{plainnat}
\bibliography{sample}

\end{document}